\documentclass[lettersize,journal]{IEEEtran}
\usepackage{amsmath,amsfonts}
\usepackage{algorithmicx}
\usepackage{algcompatible}
\usepackage{algorithm}
\usepackage{array}
\usepackage[caption=false,font=normalsize,labelfont=sf,textfont=sf]{subfig}
\usepackage{textcomp}
\usepackage{stfloats}
\usepackage{url}
\usepackage{verbatim}
\usepackage{graphicx}
\usepackage[utf8]{inputenc}
\usepackage{newunicodechar}
\newunicodechar{⁻}{\textsuperscript{-}}
\usepackage{setspace}
\usepackage{cite}
\usepackage{enumitem}
\def\BibTeX{{\rm B\kern-.05em{\sc i\kern-.025em b}\kern-.08em
    T\kern-.1667em\lower.7ex\hbox{E}\kern-.125emX}}
\begin{document}

\title{A Hybrid Deep Learning Framework for Emotion Recognition in Children with Autism During NAO Robot-Mediated Interaction}

\author{
Indranil~Bhattacharjee\textsuperscript{1},
Vartika~Narayani~Srinet\textsuperscript{2},
Anirudha~Bhattacharjee\textsuperscript{2},
Braj~Bhushan\textsuperscript{2},
Bishakh~Bhattacharya\textsuperscript{2,*} \\
\textsuperscript{1}Department of Information Technology, School of Engineering,
Cochin University of Science and Technology, Kochi, Kerala, India \\
\textsuperscript{2}Indian Institute of Technology Kanpur, Uttar Pradesh, India \\
Emails: indranil@ug.cusat.ac.in, vartikana23@iitk.ac.in, anirub@iitk.ac.in, brajb@iitk.ac.in, *bishakh@iitk.ac.in
}

\maketitle

\begin{abstract}
Understanding emotional responses in children with Autism Spectrum Disorder (ASD) during social interaction remains a critical challenge in both developmental psychology and human-robot interaction. This study presents a novel deep learning pipeline for emotion recognition in autistic children in response to a name-calling event by a humanoid robot (NAO), under controlled experimental settings. The dataset comprises of around 50,000 facial frames extracted from video recordings of 15 children with ASD. A hybrid model combining a fine-tuned ResNet-50-based Convolutional Neural Network (CNN) and a three-layer Graph Convolutional Network (GCN) trained on both visual and geometric features extracted from MediaPipe FaceMesh landmarks. Emotions were probabilistically labeled using a weighted ensemble of two models: DeepFace's and FER, each contributing to soft-label generation across seven emotion classes. Final classification leveraged a fused embedding optimized via Kullback-Leibler divergence.  The proposed method demonstrates robust performance in modeling subtle affective responses and offers significant promise for affective profiling of ASD children in clinical and therapeutic human-robot interaction contexts, as the pipeline effectively captures micro emotional cues in neurodivergent children, addressing a major gap in autism-specific HRI research. This work represents the first such large-scale, real-world dataset and pipeline from India on autism-focused emotion analysis using social robotics, contributing an essential foundation for future personalized assistive technologies.
\end{abstract}

\section{Introduction}
NAO humanoid robot developed by SoftBank Robotics, standing 58 cm tall with 25 degrees of freedom, is widely utilized in educational and therapeutic environments due to its semi-anthropomorphic appearance and programmable capabilities. Globally, NAO has been applied in diverse contexts ranging from children's education to autism interventions, yet its deployment in India remains a sparse scenario that presents a significant opportunity for strengthening socio-cognitive support through technology-enhanced methods.

Amid growing concerns that excessive screen time and digital media consumption may impact children's attentional capacities, there is increasing interest in robot-mediated interventions as proactive tools to foster engagement and learning. In children with Autism Spectrum Disorder (ASD), one of the hallmark early markers is a delayed or absent response to name-calling, a clinical indicator frequently used in diagnostic assessments. Evidence indicates that ASD children demonstrate heightened responsiveness and engagement when interacting with robotic agents \cite{Rudovic2018} positioning Socially Assistive Robots (SARs) like NAO as promising platforms for eliciting measurable socio-behavioral responses.

While response to name (RTN) paradigms have been previously explored within ASD diagnostic protocols, integration with robust, deep learning based emotion detection especially combining facial appearance and geometric landmark data have not been fully realized. Conventional approaches tend to rely on either texture-based convolutional models, which may miss subtle expressions, or landmark sequences, which fail to account for global affective context, discussed in \cite{li2018microexpression}.

To address this limitation, we propose a novel hybrid CNN–GCN architecture, named Fusion-N, capable of extracting and fusing multi-scale emotional cues from both RGB imagery and facial landmarks simultaneously, shown in Fig~1. Our pipeline leverages ensemble-derived soft labels from DeepFace's and FER models, enabling probabilistic training that effectively models emotion ambiguity and anticipates ASD-specific expression patterns. We evaluated this approach on a dataset comprising almost ~50,000 high-resolution frames obtained from 15 children with ASD during NAO-mediated RTN tasks and demonstrated its efficacy in accurately classifying nuanced emotion states, including fear and disgust, which are typically underrepresented in ASD datasets. This methodology contributes to the fields of affective computing, human-robot interaction, and computational neuro-psychology by introducing a multimodal framework for assessing emotion recognition in vulnerable developmental cohorts.

\begin{IEEEkeywords}Autism, NAO, Child-Robot Interaction, Emotion analysis, ResNet-50, GCN, Deepface, Mini-Xception, FER.\end{IEEEkeywords}

\section{Related Work}
Facial expression recognition (FER) has long been a cornerstone in affective computing and human-computer interaction. Among the most widely adopted face detection pipelines is the Multi-task Cascaded Convolutional Neural Network (MTCNN) framework by Zhang et al. \cite{zhang2016joint}, which remains a benchmark for real-time face detection and alignment due to its efficiency in bounding-box regression and landmark localization. 
\begin{figure}[h]
\centering
\includegraphics[width=3.5in]{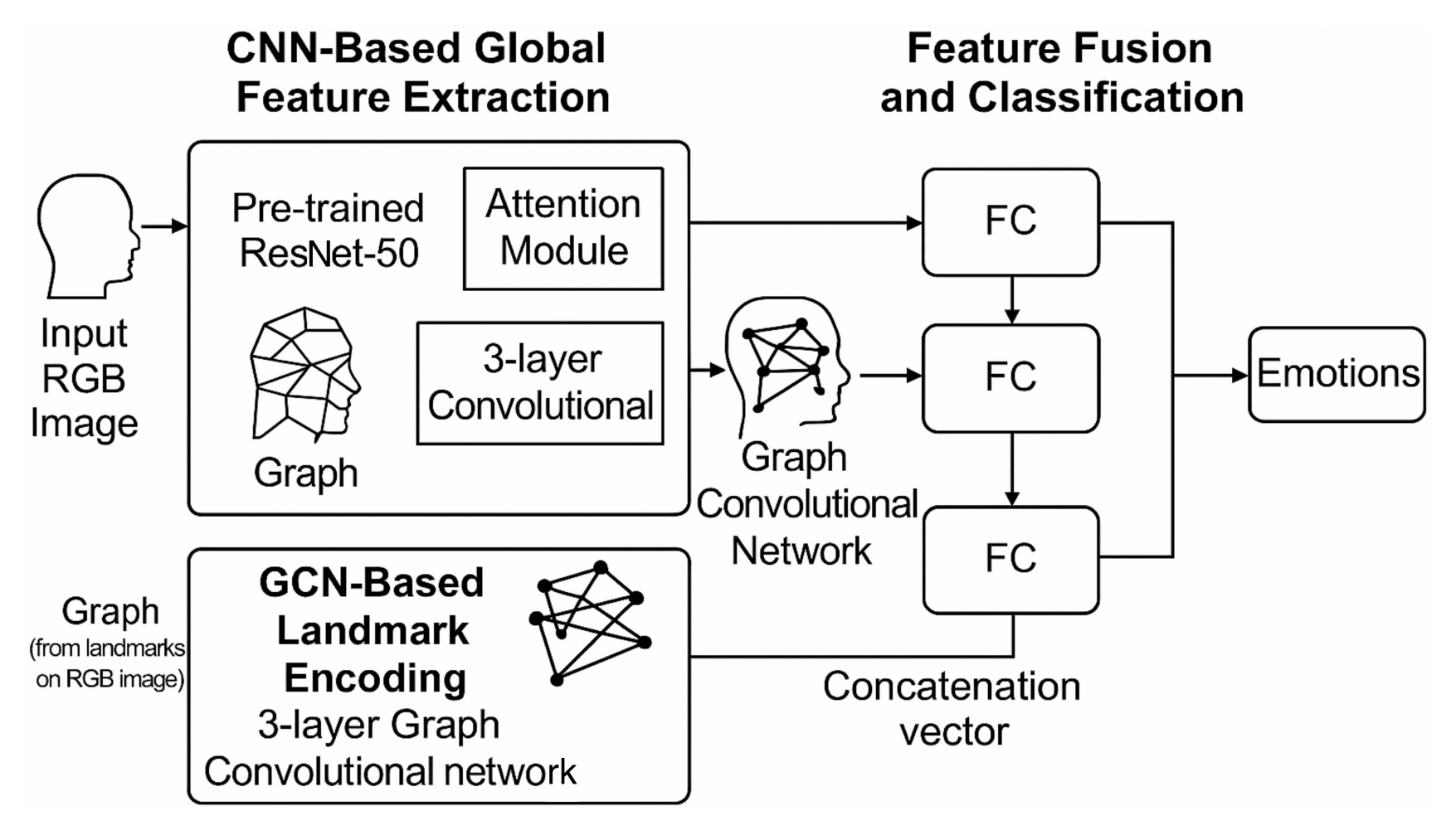}
\caption{A focused top-level view of our multimodal pipeline structure. Fusion-N, the novel hybrid framework made using ResNet-50 and GCN.}
\label{fig:unimodal_pipeline}
\end{figure}

For facial landmark extraction, Lugaresi et al.\cite{lugaresi2019mediapipe} introduced MediaPipe FaceMesh, which provides dense 3D landmark detection (468 key points), forming a strong basis for extracting geometric and relational facial features and facilitating more nuanced understanding of facial structure and microexpressions. Graph Convolutional Networks (GCNs) have become another cornerstone in modeling structured data by combining node features with graph topology. A seminal work by Kipf and Welling \cite{kipf2017gcn} introduced the modern GCN architecture, which efficiently performs semi-supervised node classification via layer-wise propagation based on graph Laplacians.
To label emotional states, researchers have increasingly moved beyond single-label supervision to probabilistic soft labels that account for ambiguity and class overlap. The DeepFace library \cite{serengil2020deepface}, with its robust backbones such as VGG‑Face \cite{parkhi2015deepface} , FaceNet \cite{schroff2015facenet} backbones, has been widely adopted for face recognition, especially in facial datasets characterized by real-world variability. Similarly, Mini-Xception architectures trained on FER2013 \cite{arriaga2017real} have demonstrated competitive performance with lower computational overhead, making them ideal for ensemble frameworks. These models are particularly helpful in analyzing common human expressions. A recent system, SENSES‑ASD\cite{abu2024senses}, utilized Mini‑Xception (trained on FER‑2013) for facial emotion recognition in autistic adults and achieved a validation accuracy of approximately 60\%\cite{abu2024senses}.
 The integration of DeepFace  (Mini-Xception) and FER-based predictions through weighted averaging forms a non-obvious soft-label calibration method which is better suited for neuro-divergent datasets where emotional ambiguity is prevalent.
 
The increasing use of GCNs has also led to hybrid models that combine image based CNN features with graph based structural information. Bin Li and Lima \cite{li2021facial} implemented a ResNet-50 based architecture for facial expression recognition, showcasing its robustness across benchmark datasets. Our model Fusion-N integrates a ResNet-50 variant for global semantic extraction and a topology-aware GCN over facial landmarks to generate spatial embeddings. This hybrid architecture demonstrates higher accuracy and better generalization, especially when analyzing subtle or masked emotions such as fear or disgust emotions that are often underrepresented and harder to detect.

While many studies have focused on emotion recognition in typical populations, relatively fewer have addressed the unique challenges posed by children with ASD. \cite{guillon2014emotion} underscored the importance of developing systems that can support or augment emotion recognition capabilities. The role of assistive technologies, particularly humanoid robots such as NAO, has grown significantly in autism research. Robins et al. \cite{robins2004robots} were among the first to demonstrate the potential of robots in engaging children with ASD through structured interactions. Rudovic et al. \cite{Rudovic2018} expanded this domain by introducing personalized machine learning algorithms that enabled robots to adapt to individual emotional patterns in children with ASD.

Studies show that NAO robot interventions have the potential to enhance emotional expressiveness and social engagement in children with ASD significantly. Robot therapy promotes communication in minimally verbal children, increases social engagement with imitation activities, and stimulates better classroom participation compared to normal settings~\cite{Feil-Seifer2011, Tapus2007, Dautenhahn2005}. This is particularly significant in name-calling tests, in which a child's reaction to their own name offers an insight into social awareness, attention, and affective states, all of which are significant diagnostic indicators in early diagnosis of autism. Costescu et al.~\cite{costescu2015comparison} similarly proved that children with ASD were more socially responsive when the NAO robot was engaged in imitative play and joint-attention exercises. These results strongly advocate for combining NAO-based interaction paradigms with computationally sophisticated emotion-analysis pipelines through the combination of soft-label supervision, dense facial-geometry modeling, and robot-mediated data collection. Such an integration provides a solid framework to study affective behavior in autistic children in ethically approved, ecologically valid experimental environments.

\begin{figure}[h]
\centering
\includegraphics[width=3 in]{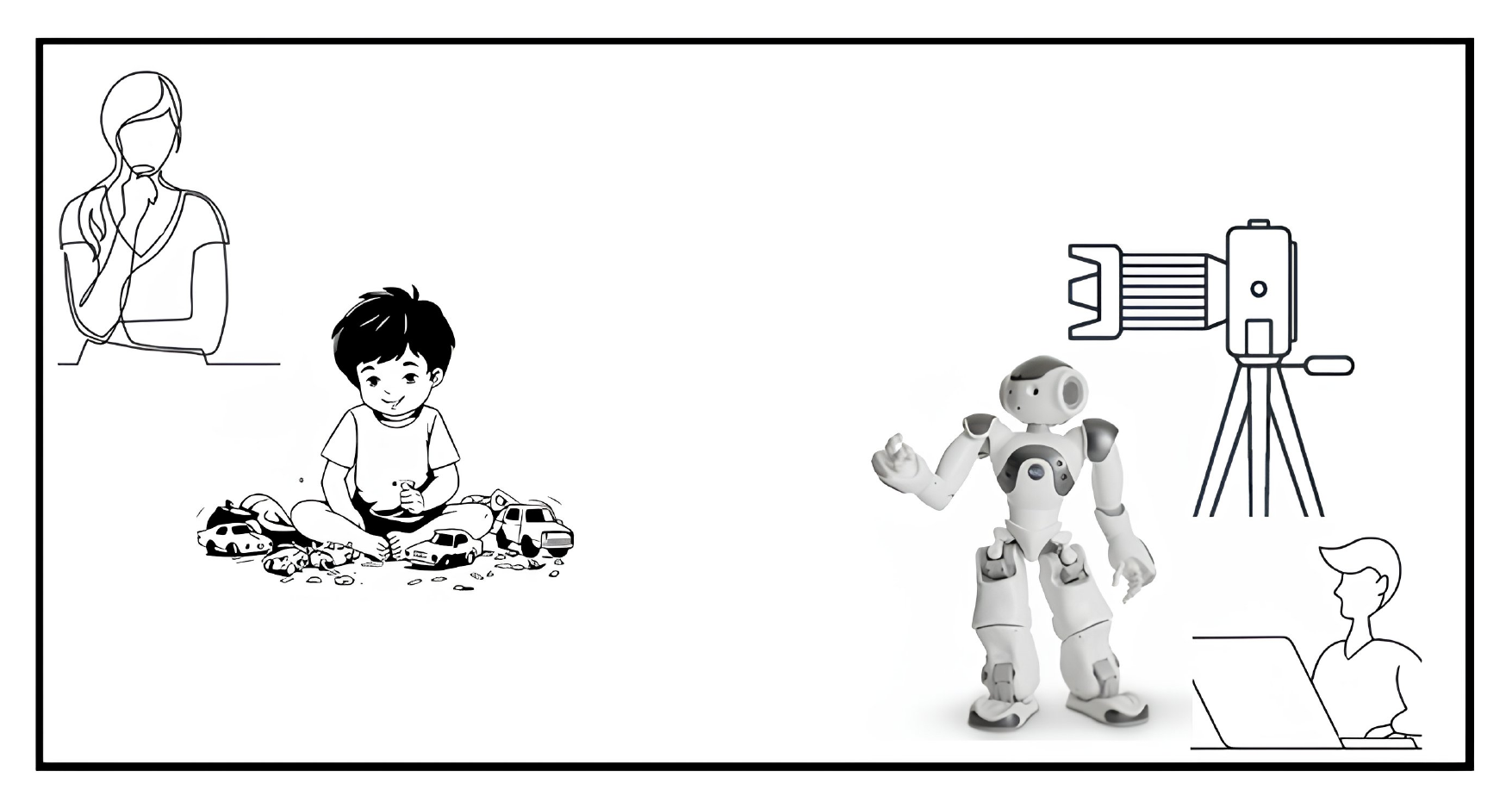}
\caption{This figure illustrates the setup of an autistic child engaging in free play in an unbiased environment with NAO and a facilitator seated nearby.  }
\label{Figure 1}
\end{figure}

\begin{table}[htbp]
\centering
\caption{Data Specifications}
\label{tab:face_preprocessing}
\renewcommand{\arraystretch}{1.2}
\begin{tabular}{|l|l|}
\hline
\textbf{Parameter} & \textbf{Value} \\
\hline
Subjects & 15 children with ASD \\
\hline
Videos & 15 (1 per child) \\
\hline
Duration & 3–5 minutes per child \\
\hline
Name Called & 12 times (randomly spread) \\
\hline
FPS for Processing & 15 \\
\hline
Frames Extracted & 48,891 \\
\hline
Label Distribution & Balanced across 7 emotions \\
\hline
\end{tabular}
\end{table}
\vspace{1em}

\section{Methodology}

The proposed emotion recognition pipeline for autistic children is a modular, multi-staged architecture designed to capture and interpret subtle affective cues from video data. The flow of controls in our pipeline is displayed in Fig. 3. The stages of this pipeline flow as follows:

\begin{figure}[t]
\centering

\includegraphics[width=3.5 in]{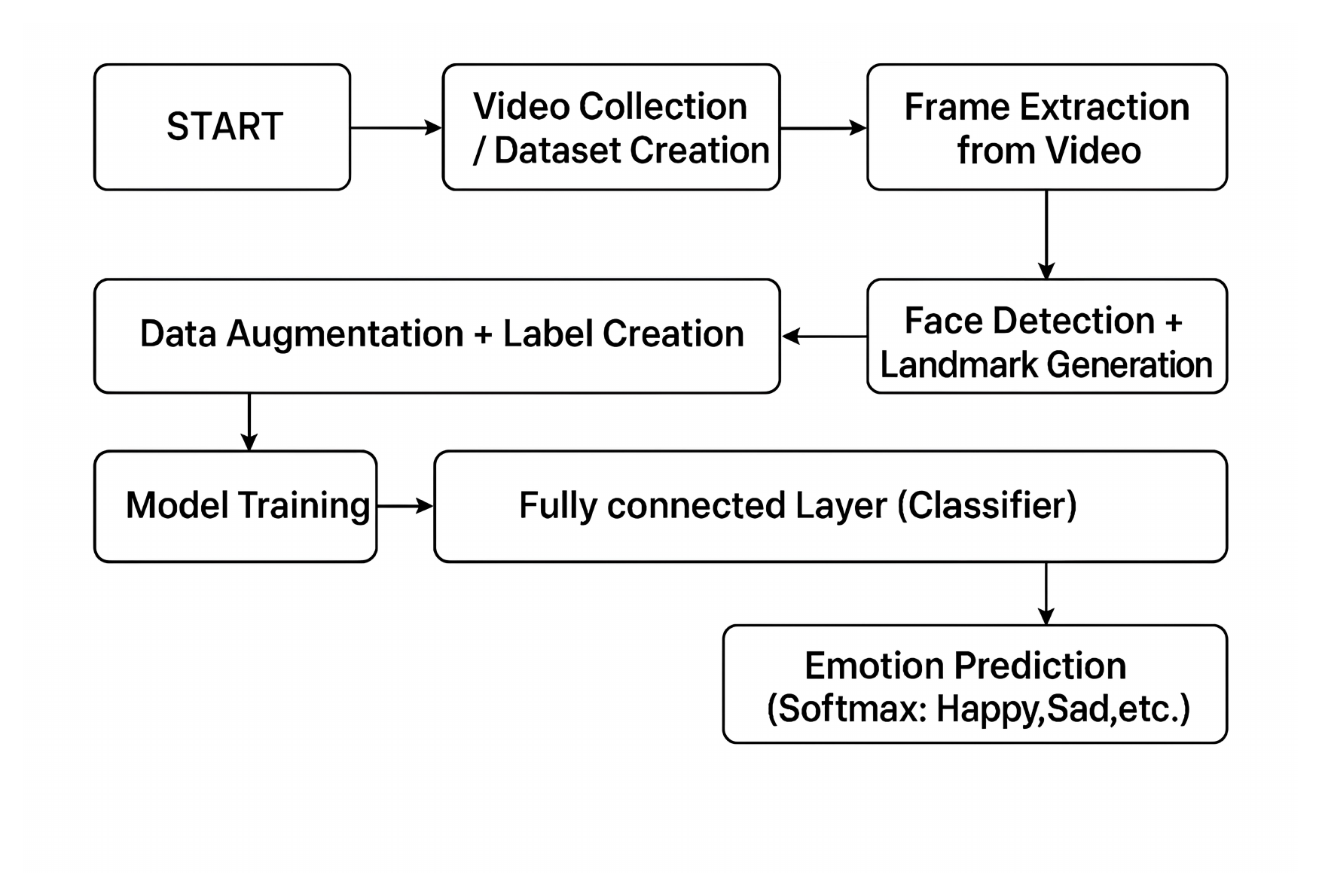}
\caption{Flowchart of the facial emotion recognition pipeline. The process begins with dataset creation through video collection, followed by face detection. Detected faces are validated, aligned, and then passed to the facial landmark extraction module. These features, along with the cropped face images, are fed into our novel hybrid model (Fusion-N) to generate emotion probability predictions.}
\label{fig:fer_pipeline}
\end{figure}

\subsection{Experimental data acquisition}

After approval of the  Institutional Ethics Committee of the Indian Institute of Technology, Kanpur  and the center head and consent from the parents, the psychological analysis report of the children was obtained to finalize our selection criterias such as studying children in mild to moderate autism spectrum and 6 to 10 years of age.

Sessions were conducted in a carefully curated environment to ensure the child's comfort, with a trusted psychologist present and strict confidentiality maintained throughout.

The child participated in a semi-structured interaction session for a duration of 3–5 minutes in a known and relaxed environment, with provision of toys and play materials to minimize stress and improve ecological validity. In this free-play setting, the NAO robot performed a pre-programmed name-calling procedure, uttering each child's name 12 times in random temporal order. The experimental configuration is shown in Fig. 2, and dataset information is given in Table I.

\subsection{Face Extraction}

Face detection is performed using the Multi-task Cascaded Convolutional Neural Network (MTCNN), which jointly handles face localization and bounding-box regression. To ensure clean inputs, frames are filtered for blur and validity, followed by secondary verification using Dlib's CNN/HOG detector (results were the same in both cases) via \texttt{face\_recognition.face\_locations}, discussed by \cite{king2017dlibface} to reduce false positives. To address MTCNN's over-cropping, temporary dynamic padding is applied during validation, though only unpadded images are retained for downstream processing. Verified bounding boxes are used to extract 468 3D facial landmarks via MediaPipe Face Mesh, capturing dense anatomical regions (e.g., brows, lips, jawline). Landmarks are normalized using min-max scaling relative to the nose tip for scale, rotation, and translation invariance. The resulting data is exported in CSV format for graph-based modeling.

\subsection{Probabilistic Soft Label Generation}

To accommodate the ambiguity of expressions common in ASD, we employed a soft-labeling mechanism using ensemble fusion. Emotion probabilities are computed by aggregating predictions from two independently trained models:

\begin{itemize}
    \item \textbf{DeepFace:} A Mini-Xception model trained on FER-2013 ~\cite{arriaga2017real}, providing semantic emotion embeddings.
    \item \textbf{FER:} A custom CNN-based model by Shenk~\cite{shenk2020fer}, also trained on FER-2013, outputting 7-class softmax distributions.
\end{itemize}

The final distribution $\mathbf{y}_{\text{final}} \in \mathbb{R}^7$ is obtained as a weighted average:

\[
\mathbf{y}_{\text{final}} = \frac{1}{3} \cdot \mathbf{y}_{\text{DeepFace}} + \frac{2}{3} \cdot \mathbf{y}_{\text{FER}}
\]

  FER is trained and tested more  on low-quality images. During our validation tests, FER consistently produced lower error rates compared to DeepFace in the low-resolution scenarios ~\cite{delovski2023emotion}. That's the reason why assigning a greater weight to FER in the ensemble enhances overall prediction quality , the ensemble is relying more on the model which is performing better under the real conditions of our data provided in the Table IV. Both models are trained on tightly cropped, aligned face images from FER-2013. Although they include their own detectors, we supplied preprocessed face crops to minimize issues such as failed detection, incorrect scale, or orientation, thereby improving prediction robustness. This ensemble strategy mitigates model-specific bias and enhances reliability across diverse visual inputs, as demonstrated in Table~II. The full soft-labeling workflow is illustrated in Fig.~5. 

\begin{table*}[ht]
\centering
\caption{Comparison of Emotion Detection Models and Fusion Strategy Used in the Proposed Pipeline}
\label{tab:emotion_model_comparison}
\renewcommand{\arraystretch}{1.2}
\begin{tabular}{|p{1.5cm}|p{2.2cm}|p{2.5cm}|p{1.5cm}|p{5.5cm}|}
\hline
\textbf{Model Source} & \textbf{Architecture} & \textbf{Output Type} & \textbf{Fusion Weight} & \textbf{Rationale} \\
\hline
\textbf{DeepFace} & Mini-Xception & 7-class probability distribution & $1/3$ & Lightweight CNN pretrained on FER-2013, efficient for real-time inference \\
\hline
\textbf{FER} & Custom CNN (\texttt{fer} library) & 7-class probability distribution & $2/3$ & Accurate and fast, empirically better on subtle emotions \\
\hline
\textbf{Ensemble Logic} & Weighted average & Final 7-class soft probabilities & -- & Reduces neutral bias using penalty regularization and sharpens predictions via temperature scaling \\
\hline
\end{tabular}
\end{table*}

\subsection{Hybrid CNN-GCN Classification (Fusion-N)}

We introduced \textit{Fusion-N}, a dual-branch architecture that jointly processes pixel-level and geometric information. A schematic diagram of Fusion-N is shown in Fig. 4.

\subsubsection{CNN Branch}

 \begin{figure}[t]
     \centering
     \includegraphics[width=0.45\textwidth, trim= 20 0 180 0, clip]{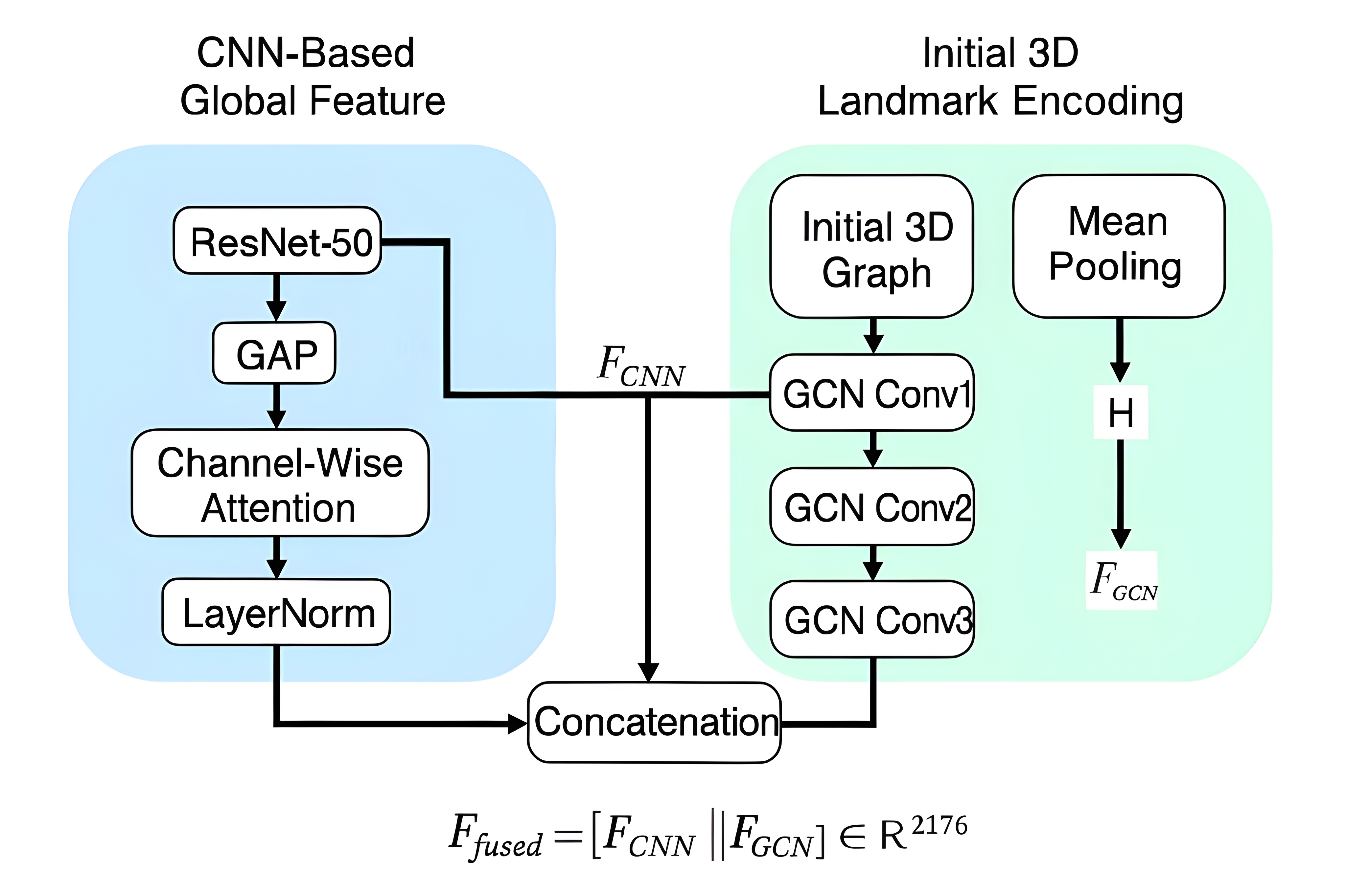}
     \caption{Simplified architecture of the proposed \textit{Fusion-N} model.
    The network consists of two parallel branches: a CNN-based global feature extractor (left) that uses ResNet-50 with channel-wise attention to produce the global descriptor $\mathbf{F}_{\text{CNN}} \in \mathbb{R}^{2048}$, and a GCN-based geometric branch (right) that encodes 3D facial landmarks into $\mathbf{F}_{\text{GCN}}$ via a stack of GCN layers and mean pooling. The two feature streams are fused via simple concatenation after intra-branch attention refinement, resulting in the final representation $\mathbf{F}_{\text{fused}} \in \mathbb{R}^{2176}$.}
     \label{Figure 3}
 \end{figure}

 \begin{figure*}[t]
     \centering
     \includegraphics[width=0.85\textwidth]{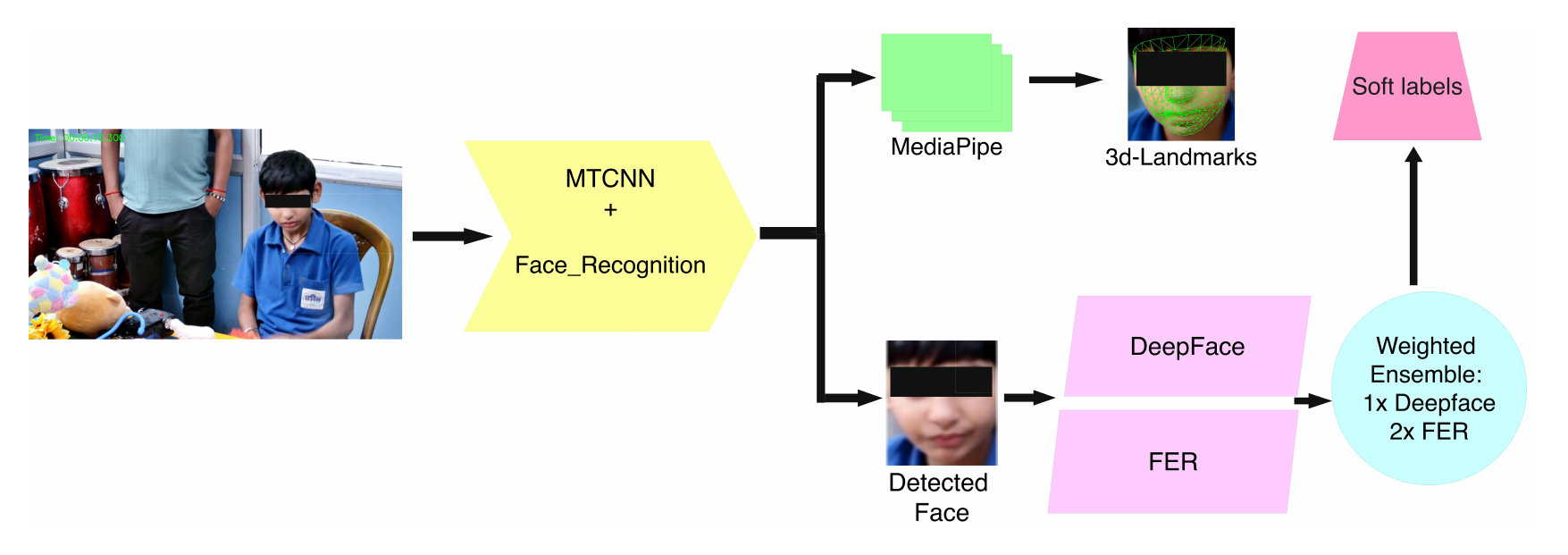}
     \caption{Segmented architecture of the pipeline, illustrating the phases of face detection using MTCNN, face validation via \texttt{face\_recognition}, landmark extraction using MediaPipe FaceMesh, and the creation of soft labels for training the Fusion-N model.}
     \label{fig:unimodal_pipeline_2}
 \end{figure*}

Aligned RGB face images of size $224 \times 224 \times 3$ are passed through a ResNet-50 backbone, with the first 44 parameters tensors frozen and the rest fine-tuned. The output feature vector $\mathbf{f}_{\text{img}} \in \mathbb{R}^{2048}$ captures global semantic information and is refined by an attention module.

\subsubsection{GCN Branch} 
Facial graphs are constructed from 468 landmarks with edges defined by facial geometry (jawline, eyebrows, eyes, mouth). 
A 3-layer Graph Convolutional Network (GCN) extracts relational features, and the pooled 128-dimensional embedding $f_{\text{geom}} \in \mathbb{R}^{128}$ is further refined with attention. 

\subsubsection{Fusion and Classification}

The concatenated feature vector $\mathbf{f}_{\text{joint}} = [\mathbf{f}_{\text{img}} \| \mathbf{f}_{\text{geom}}] \in \mathbb{R}^{2176}$ is passed through a series of dense layers with dropout and LayerNorm. Emotion class probabilities are predicted using a softmax layer.

\subsubsection{Loss Function}

Model training minimizes KL divergence between predicted scores $\mathbf{s}_\theta$ and calibrated targets $\tilde{\mathbf{y}}$:
\begin{equation}
\mathcal{L}_{\text{KL}} = \sum_i \tilde{y}_i \log\left(\frac{\tilde{y}_i}{s_{\theta,i}}\right)
\end{equation}
where $i \in \{1, \dots, C\}$ indexes emotion classes.

\subsection{Framework Used}

Face detection and pre-processing were performed using MTCNN, followed by validation through the \texttt{face\_recognition} library from DLib\cite{king2009dlib}. Quality control was implemented using Laplacian variance thresholding to remove blurry frames. Geometric normalization was applied to ensure alignment consistency.

For pose-invariant facial landmark extraction, we utilized the Face Mesh solution provided by MediaPipe \cite{lugaresi2019mediapipe}
. The 3D coordinates were normalized prior to further processing.

To generate soft emotion labels, the DeepFace \cite{serengil2020lightface} and FER ~\cite{shenk2020fer} libraries were employed. These outputs were used in conjunction with the PyTorch \texttt{Dataset} API to structure a triplet input pipeline consisting of face images, landmarks, and corresponding soft labels.

\section{Optimization and Training Framework}

Training is done with the AdamW optimizer~\cite{loshchilov2019decoupled}, using discriminative learning rates of $3 \times 10^{-6}$ and $1 \times 10^{-5}$ for the pretrained CNN backbone and classifier head, respectively, with a global $L_{2}$ weight decay of $5 \times 10^{-4}$ to prevent overfitting~\cite{krogh1992simple}. The main criterion is the label-smoothed KL divergence (smoothing factor $=0.1$), ensuring robust learning with softened target distributions. Training stability is maintained through gradient clipping (L2 norm limit $=1.0$), while effective exploration of the loss landscape is facilitated by a cosine annealing learning rate schedule with warm restarts ($T_{0}=10$, $T_{m}=2$, $\eta_{\min}=1 \times 10^{-5}$). The evaluation metrics include per-class precision, recall, F1 score, and overall accuracy, following recommended practices for balanced and robust evaluation, especially in the presence of minority classes~\cite{powers2020evaluation}.

\section{Techniques Involved}
This section presents a detailed computational framework for multimodal emotion recognition specifically designed for subjects with Autism Spectrum Disorder (ASD). 

\subsection{Hierarchical Facial Region-of-Interest Detection}
To achieve precise anatomical localization of facial regions, we implemented a dual-step face verification strategy. Initially, the Multi-task Cascaded Convolutional Networks (MTCNN) was employed. This preliminary detector helped localize potential facial regions.

To ensure high-quality face inputs, all images were first filtered for blur (Laplacian threshold = 25) and low-confidence detections (MTCNN score $<$ 70\%). A secondary validation using Dlib’s \texttt{face\_recognition} (CNN/HOG) filtered out non-facial or corrupted frames; both backends yielded comparable results with only clean, centered faces retained. Faces smaller than 30$\times$30 were discarded, and accepted crops were resized to 224$\times$224.

To correct MTCNN's tight cropping, temporary padding was applied during verification (not saved), preserving undistorted facial features. Final verified crops were aligned using reused MTCNN boxes and forwarded for landmark detection.
Later, MediaPipe Face Mesh extracted 468 normalized 3D landmarks per face, enabling pose-invariant, topology-aware CSV features for robust graph modeling of neurodivergent expressions.

\subsection{Confidence-Calibrated Label Incorporation}
Several interactive facial emotion recognition tools targeting autistic individuals have been proposed. For instance, Abu‑Nowar et al. (2024) introduced SENSES‑ASD a web/mobile platform utilizing a compact Mini‑Xception CNN (~60K parameters) trained on FER‑2013 (35,887 grayscale images across seven emotions). The system initially achieved ~60\% validation accuracy, which improved to ~66\% after tuning, with training accuracy reaching ~71\%~\cite{abu2024senses}.
To account for the semantic ambiguity and inter-class overlap prevalent in ASD expression datasets, we proposed a confidence-aware  novel soft-labeling mechanism based on ensemble modeling. This approach jointly leverages the high representational capacity of DeepFace (Mini-Xception) and the robustness of FER network.

\subsubsection*{Dual-Model Ensemble}

\paragraph{DeepFace Backbone}
We used the Mini-Xception model from DeepFace~\cite{arriaga2017real}, a lightweight CNN trained on FER-2013, producing softmax outputs $\mathbf{p}_{\text{DF}} \in \Delta^C$ across $C=7$ emotion classes. These predictions contribute to our ensemble fusion strategy. Despite its efficiency, Mini-Xception has shown performance comparable to human-level accuracy on benchmark datasets.

\paragraph{FER Supplement}
To enhance robustness against occlusions and low-resolution inputs, we incorporate a parallel FER branch (Shenk~\cite{shenk2020fer}) via the \texttt{fer} library. It outputs $\mathbf{p}_{\text{FER}} \in \Delta^C$, also trained on FER-2013 but using a deeper CNN than Mini-Xception.

\paragraph{Weighted Fusion}
The final ensemble prediction is computed as:
\begin{equation}
\mathbf{p}_{\text{ens}} = \frac{2}{3} \cdot \mathbf{p}_{\text{FER}} + \frac{1}{3} \cdot \mathbf{p}_{\text{DF}}
\end{equation}
Emotion classifiers often over-predict the \textit{neutral} class. 
To mitigate this bias, we apply a multiplicative penalty:
\begin{equation}
    \tilde{p}_{\text{neutral}} = \gamma \cdot p_{\text{fuse,neutral}}, 
    \quad \gamma = 0.7,
\end{equation}
where $p_{\text{fuse}}$ denotes the fused distribution over emotion classes 
and $\gamma$ is a clinically validated scaling factor.
The adjusted vector $\tilde{p}$ is re-normalized to ensure a valid probability distribution:
\begin{equation}
    \hat{p} = \text{softmax}(\tilde{p}).
\end{equation}
Here, $\hat{p}$ represents the probability distribution across emotion 
classes after neutral adjustment.

Temperature scaling ($T = 0.7$) is applied via \texttt{np.power(final\_vector, 1.0/T)} followed by normalization, enhancing distribution sharpness. This fusion balances speed and sensitivity. Mini-Xception favors real-time applications, while FER shows improved response to subtle expressions.

\subsection{Primary Model Architecture: Fusion-N}
We introduced Fusion-N, a hybrid deep neural network combining Convolutional Neural Network (a fine-tuned ResNet-50) and Graph Convolutional (GCN) to integrate global appearance features and localized relational (landmark) geometry. The architecture of Fusion-N is shown in Fig.~6.

\subsection*{a. Attention on CNN feature vector}

\begin{equation}
\mathbf{F}_{\text{CNN}}^{\mathrm{attn}} = \mathbf{A}_{\text{CNN}} \odot \mathbf{F}_{\text{CNN}}
\end{equation}

where $\odot$ denotes the element-wise (Hadamard) product \cite{wiki:hadamard,holt2013elementwise}, $\mathbf{A}_{\text{CNN}}$  and $\mathbf{F}_{\text{CNN}}^{\text{attn}}$ is the refined CNN feature vector used downstream.

\vspace{1em}

\subsection*{b. Aggregated GCN Features}

\begin{equation}
\mathbf{F}_{\text{GCN}} = \frac{1}{N} \sum_{i=1}^{N} \mathbf{H}_i^{(3)}
\end{equation}

    $\mathbf{F}_{\text{GCN}}$ denotes the aggregated node representation after three GCN layers,
    $\mathbf{H}_i^{(3)}$ is the output node features from the third GCN layer for the $i^\text{th}$ node and
    $N$ represents number of nodes (e.g., facial landmarks).
    $\sum_{i=1}^{N} \mathbf{H}_i^{(3)}$ is the mean (or sum) of the output features from all nodes in the third GCN layer.
  
This summarizes GCN features by aggregating the landmark node embeddings after the third GCN layer and mean pooling creates a single global feature vector per face.

\subsection*{c. Feature Fusion}

\begin{equation}
\mathbf{F}_{\text{fused}} = \left[ \mathbf{F}^{\text{attn}}_{\text{CNN}} \, || \, \mathbf{F}_{\text{GCN}} \right]
\end{equation}

where, $\mathbf{F}_{\text{fused}}$ is the final fused feature representation obtained by concatenating $\mathbf{F}_{\text{CNN}}$ (attention-weighted CNN features) and $\mathbf{F}_{\text{GCN}}$ (aggregated GCN features), denoted by the concatenation operator $\left[ \,\Vert\, \right]$.

This equation explains the concatenation of the features extracted from CNN (with channel-wise attention) and GCN to form a unified representation that combines both appearance and geometric information, and this  \textbf{fused vector} is forwarded to the classification head.
\\
\subsubsection{ CNN-Based Global Feature Extraction}
We leverage a pre-trained ResNet-50 backbone. ResNet-50 backbone extracts high-level features from facial images, incorporates residual learning through skip connections.
We used the standard ResNet-50 architecture \cite{he2016deep}, comprising four residual stages with bottleneck blocks. The original ResNet-50 uses Batch Normalization, ReLU activations, and identity skip connections within its residual blocks to facilitate residual learning. However, in our architecture, we additionally apply a Layer Normalization step after the attention module to stabilize the reweighted feature distribution before fusion with the GCN branch.
 The final FC layer is removed, and the rest of the network is retained up to the Global Average Pooling (GAP) layer. This transforms ResNet-50 into a strict feature extractor, with the GAP layer producing a 2048-dimensional feature vector for each input image.

 We adopt partial fine-tuning by specifically freezing first 44 parameter tensors while the remaining tensors are fine-tuned, which enable learning domain-specific features relevant to autism-oriented emotion data.

To further enhance the discriminative capacity of the extracted features, a lightweight attention module is appended after ResNet-50. This module comprises two fully connected layers with ReLU and Sigmoid activations. The resulting output is a learned attention weight vector that reweights the 2048-dimensional features, emphasizing the most informative components.

The feature map $\mathbf{F}_{\text{CNN}} \in \mathbb{R}^{2048}$ is refined using an attention module applied on the feature vector:
\begin{equation}
\mathbf{A}_{\text{CNN}} = \sigma(W_2 \cdot \text{ReLU}(W_1 \cdot \mathbf{F}_{\text{CNN}})) \quad 
\end{equation}
Here, $\mathbf{F}_{\text{CNN}}$ is the 2048‑dimensional raw feature vector from the last ResNet layer, $W_1$ and $W_2$ are learned fully‑connected weight matrices, ReLU is the rectified linear activation, $\sigma$ is the element‑wise sigmoid function (squeezing values to [0,1]), and $\mathbf{A}_{\text{CNN}}$ is the attention weight vector (the same size as $\mathbf{F}_{\text{CNN}}$).
\\

\subsubsection{ GCN-Based Landmark Encoding}
We represent each face as a fixed-topology graph $\mathcal{G} = (V, E)$ where $|V| = 468$, and edges are manually constructed based on facial geometry (jawline, eyebrows, eyes, and mouth), partially following the the MediaPipe topology (i.e., edge-index). A 3-layer GCN computes node embeddings:
\begin{equation}
\mathbf{H}^{(l+1)} = \text{ReLU}(\text{GCNConv}(\mathbf{H}^{(l)}, E)), \quad \mathbf{H}^{(0)} = X
\end{equation}
Here, $\mathbf{H}^{(\ell)}$ is the node-feature matrix output by layer $\ell$, $E$ represents the graph’s edge list or adjacency matrix, and the $\text{GCNConv}$ operator, originating from Kipf and Welling’s seminal GCN model \cite{kipf2016gcn} and implemented in PyTorch Geometric \cite{pyg_gcnconv} performs the graph convolution. $X$ is the initial $468 \times 3$ matrix of landmark coordinates. ReLU activation is applied in the first two GCN layer, while the third produces the final 128-D embeddings.
\begin{figure*}[t]
  \centering
  \includegraphics[width=0.95\textwidth]{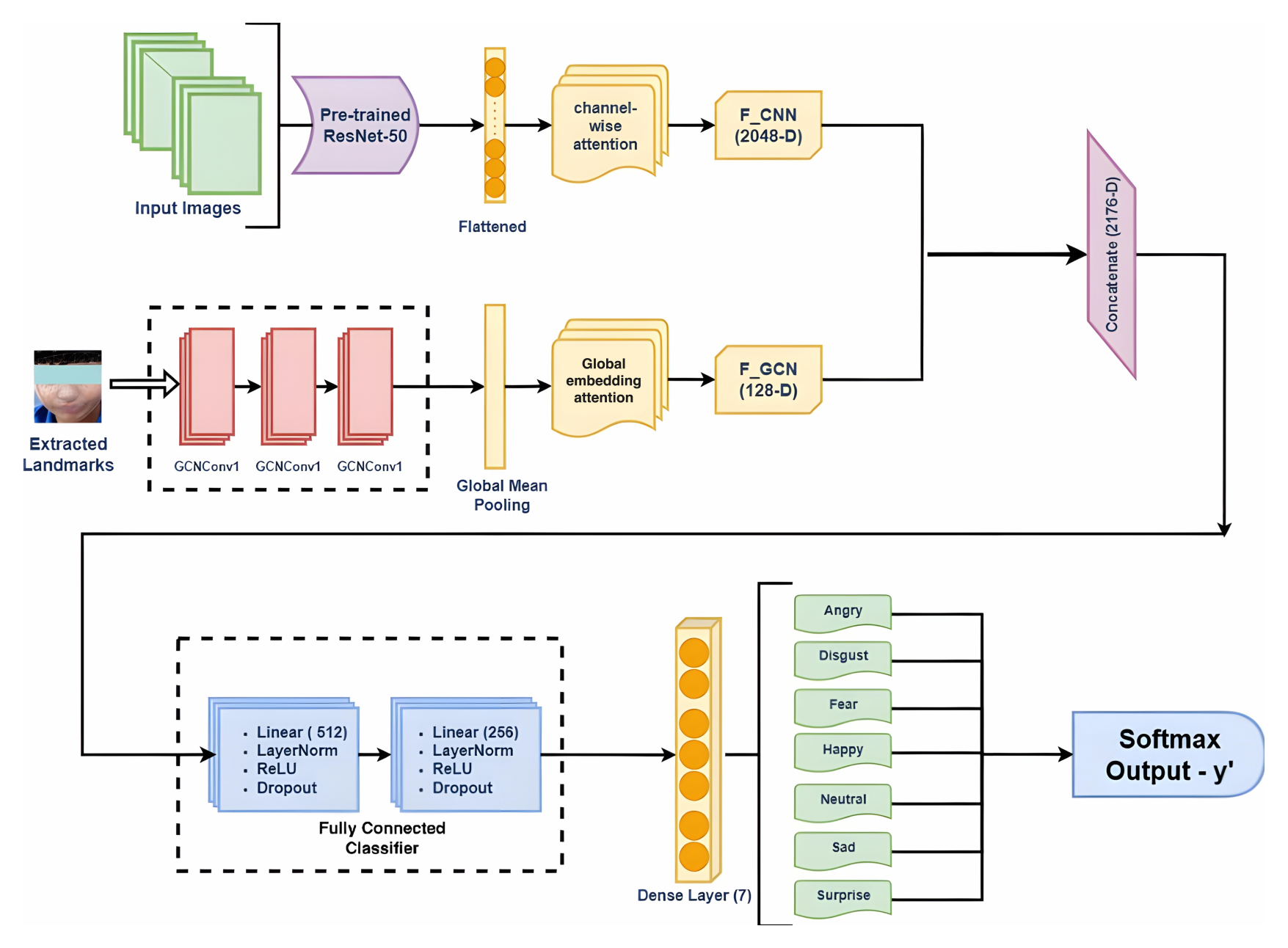}
  \caption{Architecture of the proposed Fusion-N model for facial emotion recognition. The framework comprises two branches: (i) a global feature extractor using a pre-trained ResNet-50 with an attention module applied on the 2048-D feature vector($F_{\text{CNN}}$), and (ii) a geometric branch processing 3D facial landmarks through stacked GCN layers with mean pooling, followed by an attention module to refine the global landmark embedding ($F_{\text{GCN}}$). The features are fused via concatenation, forming a joint descriptor passed through fully connected layers with layer normalization, ReLU activation, and dropout. The final dense layer outputs emotion class probabilities using softmax activation.}
  \label{fig:wide_image}
\end{figure*}

Stacking the 3 GCN layers enables each landmark to gather information from its neighbors and neighbors-of-neighbors. A \texttt{try-except} block is implemented to handle cases where the GCN fails. In such cases, a zero vector of dimension-128 is filled in to maintain consistency.

Mean-pooled, then attention-refined yields:
\begin{equation}
\mathbf{F}_{\text{GCN}} = \text{Attn}\left(\frac{1}{N} \sum_{i=1}^{N} \mathbf{H}_i^{(3)}\right)
\end{equation}

Here, $\mathbf{H}_i^{(3)}$ denotes the 128‑D embedding of landmark $i$ after three GCN layers, $\mathrm{Attn}(\cdot)$ is a small fully‑connected attention module applied on the pooled global embedding and $N$ is the total number of landmarks (468). Layer Normalization is applied prior fusion.

\subsubsection{Feature Fusion and Classification}

While CNN and GCN features are concatenated for representational purposes, the fused representation $\left[ \mathbf{F}^{\text{attn}}_{\text{CNN}} \, \| \, \mathbf{F}_{\text{GCN}} \right] \in \mathbb{R}^{2176}$ is passed through the classification head. Both the CNN and GCN branches contribute to the final prediction.

\begin{equation}
\mathbf{F}_{\text{fused}} = [\mathbf{F}_{\text{CNN}}^{\text{attn}} \parallel \mathbf{F}_{\text{GCN}}] \in \mathbb{R}^{2176}
\end{equation}

\begin{align}
\mathbf{h}_1 &= \text{ReLU}(\text{LN}(W_1 \cdot \mathbf{F}_{\text{fused}})) \\
\mathbf{h}_2 &= \text{ReLU}(\text{LN}(W_2 \cdot \mathbf{h}_1)) \\
\hat{\mathbf{y}} &= \text{Softmax}(W_3 \cdot \mathbf{h}_2)
\end{align}

\begin{figure}[h]
  \centering
  \includegraphics[width=3.5in]{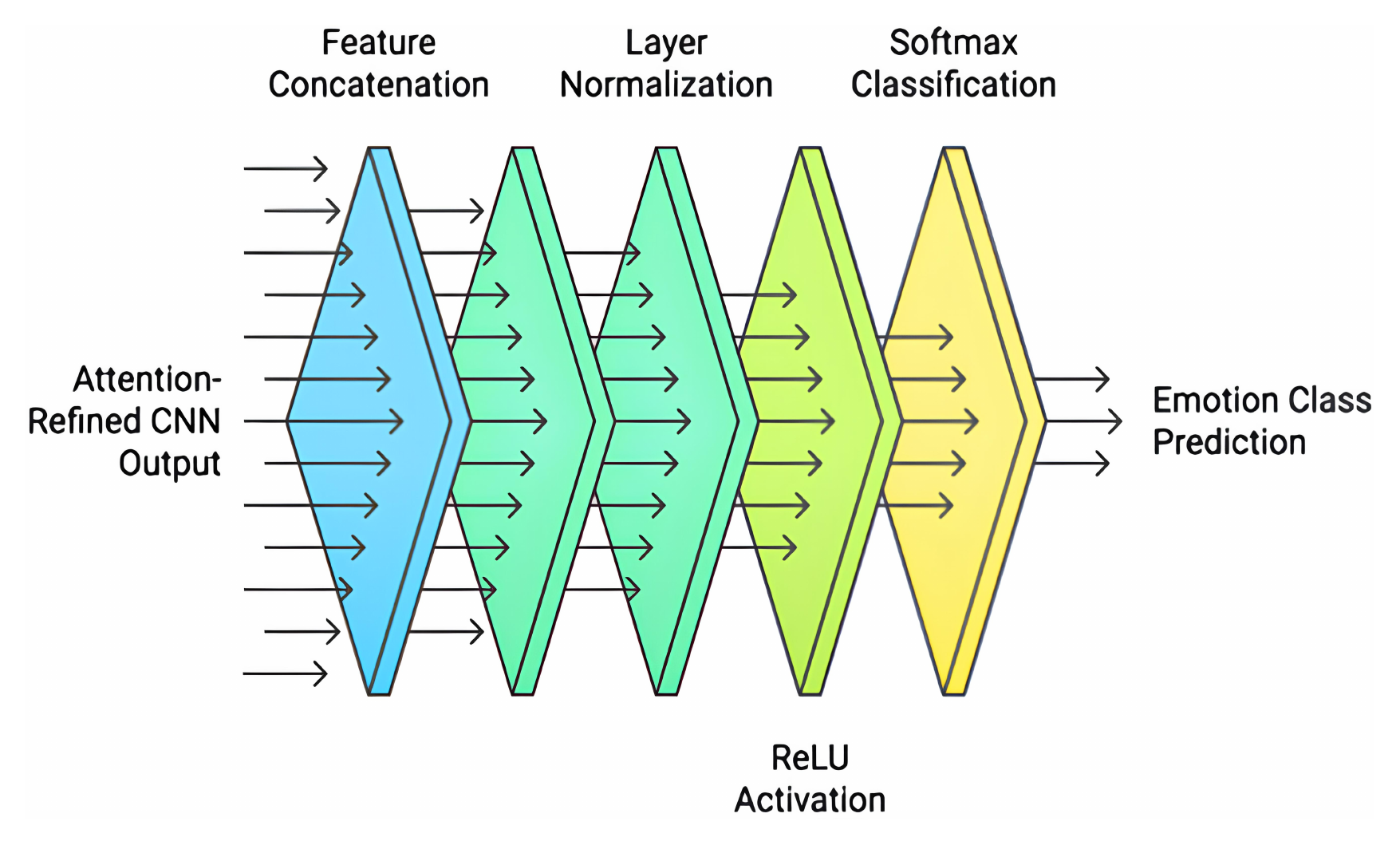}
  \caption{The attention-refined CNN feature vector (2048-D) is concatenated with the pooled GCN embedding (128-D) to get a merged 2176-D fused representation. It is passed through a classification head that contains two fully connected layers, each preceded by layer normalization, ReLU activation, and dropout for regularization. The last dense layer outputs to the target number of emotion classes, generating logits, which are then transformed into predicted class probabilities with a softmax function. This combination approach successfully combines global appearance features of the CNN and localized geometric cues of the GCN for robust facial emotion recognition.}
  \label{fig:ffcp}
\end{figure}

Here, $W_1 \in \mathbb{R}^{512\times2176}$ and $W_2 \in \mathbb{R}^{256\times512}$ are learned weight matrices, $W_3 \in \mathbb{R}^{7\times256}$ is the final linear projection, $\mathbf{h}_1$ and $\mathbf{h}_2$ are intermediate 512-dimensonal and 256-dimensional hidden vectors, respectively. ReLU is the rectified-linear activation function, LN denotes layer normalization as introduced by Ba et al.~\cite{ba2016layernorm}, $\mathbf{F}^{\text{attn}}_{\text{CNN}}$ is the 2048-dimensional attention-refined CNN feature vector and \(\hat{\mathbf{y}}\) is the predicted probability vector for seven emotion classes. Both CNN and GCN branches contribute complementary information to the fused representation.
This process has been illustrated in Fig.~\ref{fig:ffcp}.

\vspace{1em}
\noindent Inputs of Fusion-N:

\begin{enumerate}[label=\textit{\alph*)}, leftmargin=2.2em, itemindent=1em, itemsep=0.4em]
    \item Images of shape $[B, 3, H, W]$, where $B$ is the batch size, 
    3 refers to RGB channels, and $H \times W$ is the spatial resolution.

    \item Landmarks of shape $[B, 468, 3]$, where $B$ is the batch size,
    468 is the number of landmarks (from MediaPipe Face Mesh), and 
    3 denotes $(x, y, z)$ coordinates.
\end{enumerate}

\noindent{Output of Fusion-N:} Logits of shape $[B, \text{num\_classes}]$, i.e., raw scores before softmax.

\vspace{0.5em}
\noindent{Feature dimensions:} The model computes a 2048-dimensional attention-refined CNN feature vector and a 128-dimensional GCN embedding. CNN and GCN features are concatenated, and the fused 2176-dimensional vector is passed through the classification head for final emotion prediction.

\begin{algorithm}[h]
\caption{Classifier Head Pseudo‑Algorithm}
\label{alg:classifier_head_corrected}
\begin{algorithmic}[1]
\REQUIRE
  $\mathbf{X} \in \mathbb{R}^{B \times 2176}$ Fused feature matrix (batch size $B$)
  $\mathbf{W}_1 \in \mathbb{R}^{2176 \times 512},\ \mathbf{b}_1 \in \mathbb{R}^{512}$  
  $\mathbf{W}_2 \in \mathbb{R}^{512 \times 256},\ \mathbf{b}_2 \in \mathbb{R}^{256}$  
  $\mathbf{W}_3 \in \mathbb{R}^{256 \times 7},\ \mathbf{b}_3 \in \mathbb{R}^{7}$
\ENSURE
  $\mathbf{logits} \in \mathbb{R}^{B \times 7}$ Pre-softmax scores for each emotion class
\FOR{$i \leftarrow 1$ to $B$}
  \STATE \textbf{FC1:} $\mathbf{Z}_1 \gets \mathbf{X}[i] \mathbf{W}_1 + \mathbf{b}_1$  
  \STATE \textbf{LN1:} $\mathbf{N}_1 \gets \mathrm{LayerNorm}(\mathbf{Z}_1)$  
  \STATE \textbf{ReLU1:} $\mathbf{A}_1 \gets \mathrm{ReLU}(\mathbf{N}_1)$  
  \STATE \textbf{Drop1:} $\mathbf{D}_1 \gets \mathrm{Dropout}(\mathbf{A}_1,\ p=0.325)$  
  \STATE \textbf{FC2:} $\mathbf{Z}_2 \gets \mathbf{D}_1 \mathbf{W}_2 + \mathbf{b}_2$  
  \STATE \textbf{LN2:} $\mathbf{N}_2 \gets \mathrm{LayerNorm}(\mathbf{Z}_2)$  
  \STATE \textbf{ReLU2:} $\mathbf{A}_2 \gets \mathrm{ReLU}(\mathbf{N}_2)$  
  \STATE \textbf{Drop2:} $\mathbf{D}_2 \gets \mathrm{Dropout}(\mathbf{A}_2,\ p=0.275)$  
  \STATE \textbf{FC3:} $\mathbf{logits}[i] \gets \mathbf{D}_2 \mathbf{W}_3 + \mathbf{b}_3$  
\ENDFOR
\end{algorithmic}
\end{algorithm}

\subsubsection{Rationale for Hybridization}
While CNNs excel at modeling texture and color, they fail to capture geometric expressiveness, especially in ambiguous or flattened affect. GCNs, while geometrically robust, miss texture semantics. Fusion-N effectively combines both modalities, enhancing generalizability and interpretability in real-world ASD settings.

\begin{table}[htbp]
\centering
\caption{Fusion-N Architecture Comparison}
\renewcommand{\arraystretch}{1.4}
\begin{tabular}{|p{2cm}|p{2.2cm}|p{2.2cm}|}
\hline
\textbf{Characteristic} & \textbf{CNN} & \textbf{GCN} \\
\hline
Input & RGB facial images & Facial landmarks as a graph \\
\hline
Backbone & Pre-trained ResNet-50 & 3-layer Graph Convolutional Network \\
\hline
Feature Representation & Deep feature representation ($F_{\text{CNN}}$) & Graph representation ($H^{(3)}$) \\
\hline
Attention Module & Channel-wise attention & Attention after mean-pooling \\
\hline
Output Dimension & $F_{\text{CNN\_attn}} \in \mathbb{R}^{2048}$ & $F_{\text{GCN}} \in \mathbb{R}^{128}$ \\
\hline
\end{tabular}
\end{table}

\section{Results}
\subsection{Performance Comparison with Prior Work}
\subsubsection{Soft Label Generation via Ensemble Prediction}

To validate our ensemble-based emotion labeling framework for ASD contexts, we used an external dataset of autistic children curated by Dr.~Fatma~M.~Talaat~\cite{talaat2023dataset}. A representative subset of 100 images was selected with regards to maintaining a balance between the emotions and to match our cohort’s age and maximize ethnic diversity, reflecting the cross-cultural variance emphasized in~\cite{rhue2021racial,fan2023addressing}.

Each image was annotated by a licensed clinical psychologist after which 61 total images were finally analysed (some were removed on the account of the image being a little difficult to label as per and to avoid confusions) and compared against predictions from our ensemble fusion pipeline, which integrates multiple pre-trained models. The approach achieved 90.16\% accuracy relative to expert labels, demonstrating high reliability and reducing the annotation burden typical in ASD datasets.\\
Compared to DeepFace(Mini-Xception) (67.07\%), FER (71.95\%), and their average-fused variant (73.17\%), our ensemble showed superior accuracy shown in Table~IV, reinforcing its robustness and suitability for real-world clinical deployment.

\vspace{1em}
\begin{table}[h!]
\centering
\renewcommand{\arraystretch}{1.3} 
\caption{Accuracy comparison of individual models and ensemble methods.}
\label{tab:accuracy_summary}
\begin{tabular}{|p{6cm}|c|}
\hline
\textbf{Model} & \textbf{Accuracy (\%)} \\
\hline
DeepFace only & 67.07 \\
\hline
Mini-Xception (FER) & 71.95 \\
\hline
Average Fusion (DF + FER) & 73.17 \\
\hline
\textbf{Ensemble Method (Weighted Average)} & \textbf{90.16} \\
\hline
\end{tabular}
\end{table}

\subsubsection{ Hybrid Model Training and Optimization}
Several prior works have explored emotion recognition models tailored for autistic children. Alhakbani~\cite{alhakbani2024} developed a CNN trained on ASD facial images across five emotion classes, achieving 75\% accuracy, reflecting the challenges of affect recognition in this population. Smitha and Vinod~\cite{smitha2015} proposed a PCA-based system deployed on FPGA; though it reached 94.1\% on JAFFE, performance dropped to 82.3\% on real-world ASD data, underscoring domain-specific limitations. Wang et al.~\cite{wang2025} introduced a multimodal CVT architecture combining facial and speech inputs, where the facial-only branch achieved 79.12\% and the fused model reached 90\%, highlighting the benefits of cross-modal integration.

These unimodal facial expression systems (75\%, 82.3\%, 79.12\%) offer directly comparable baselines to evaluate our model, as summarized in Table~V. In contrast, our architecture built on ResNet-50 and GCN backbones was trained exclusively on an in-house ASD-specific dataset and achieved 96.2\% accuracy. This improvement demonstrates the advantage of residual feature fusion for capturing subtle affective cues often missed by traditional CNNs or hand-crafted methods.

\begin{table}[h!]
\centering
\renewcommand{\arraystretch}{1.3}
\caption{Comparison of unimodal facial-expression models evaluated on ASD datasets and their limitations.}
\label{tab:asd_models}
\begin{tabular}{|p{2.8cm}|p{1.4cm}|p{3.1cm}|}
\hline
\textbf{Study} & \textbf{Accuracy (\%)} & \textbf{Limitations} \\
\hline
Alhakbani (2024)~\cite{alhakbani2024} & $\sim$75.0 & Small and demographically narrow dataset with limited generalization. \\
\hline
Smitha \& Vinod (2015)~\cite{smitha2015} & 82.3 & Low-resolution PCA features that lacks geometric cues and real-time support. \\
\hline
Wang et al.\ (2025)~\cite{wang2025} & 79.1 & Confusion in similar emotions; no temporal modeling or ablation. \\
\hline
Our Model (2025) & 96.2 & Not real-time; possible latency in live deployment. \\
\hline
\end{tabular}
\end{table}

\subsection{Experimental results}

\subsubsection{Face pre-processing outcomes}

Our preprocessing component analyzed 48,891 frames from NAO-mediated child--robot interaction videos, recorded in a naturalistic, unconstrained environment without head fixation or behavioral restrictions. Of these, 1,600 were discarded due to blurriness and 20,170 due to missed detections, leaving 19,322 valid face crops obtained through our two-stage pipeline, corresponding to a 39.5\% face detection success rate. The comparatively low yield is consistent with the free-play setup, in which the NAO robot called the child's name 12 times across sessions involving toys and spontaneous movement. The total preprocessing duration was 40,453.52 seconds ($\approx$\,11.2~hours). A summary of these statistics is provided in Table~VI.

\begin{table}[h!]
\centering
\renewcommand{\arraystretch}{1.3} 
\caption{Summary of face preprocessing statistics}
\label{tab:face_preprocessing}
\begin{tabular}{|p{6cm}|c|}
\hline
\textbf{Metric} & \textbf{Value} \\
\hline
Total images found & 48,891 \\
\hline
Valid images & 48,886 \\
\hline
Blurry images skipped & 1,600 \\
\hline
Images with no faces & 20,170 \\
\hline
Total faces extracted & 19,322 \\
\hline
Success rate & 39.5\% \\
\hline
Processing time (seconds) & 40,453.52 \\
\hline
\end{tabular}
\end{table}

\subsubsection{Emotion distributed throughout the experiment}
Each child participated in a 200-second interaction session, with video recorded at 15 frames per second, yielding a high number of frames per participant. These were processed through our facial landmark extraction and hybrid deep learning classification pipeline.

Fig.~8 presents the distribution of emotion labels obtained via our weighted ensemble method. Most frames were classified as \textit{neutral} (8{,}969) and \textit{happy} (5{,}309), suggesting a predominance of non-negative affective states during the interaction. Moderate representation was observed for \textit{angry} (1{,}822), \textit{surprise} (1{,}605), and \textit{sad} (1{,}386), while \textit{disgust} (152) and \textit{fear} (79) were rare, likely due to the controlled experimental setting.

\begin{figure}[H]
\centering
\includegraphics[width=3.5in]{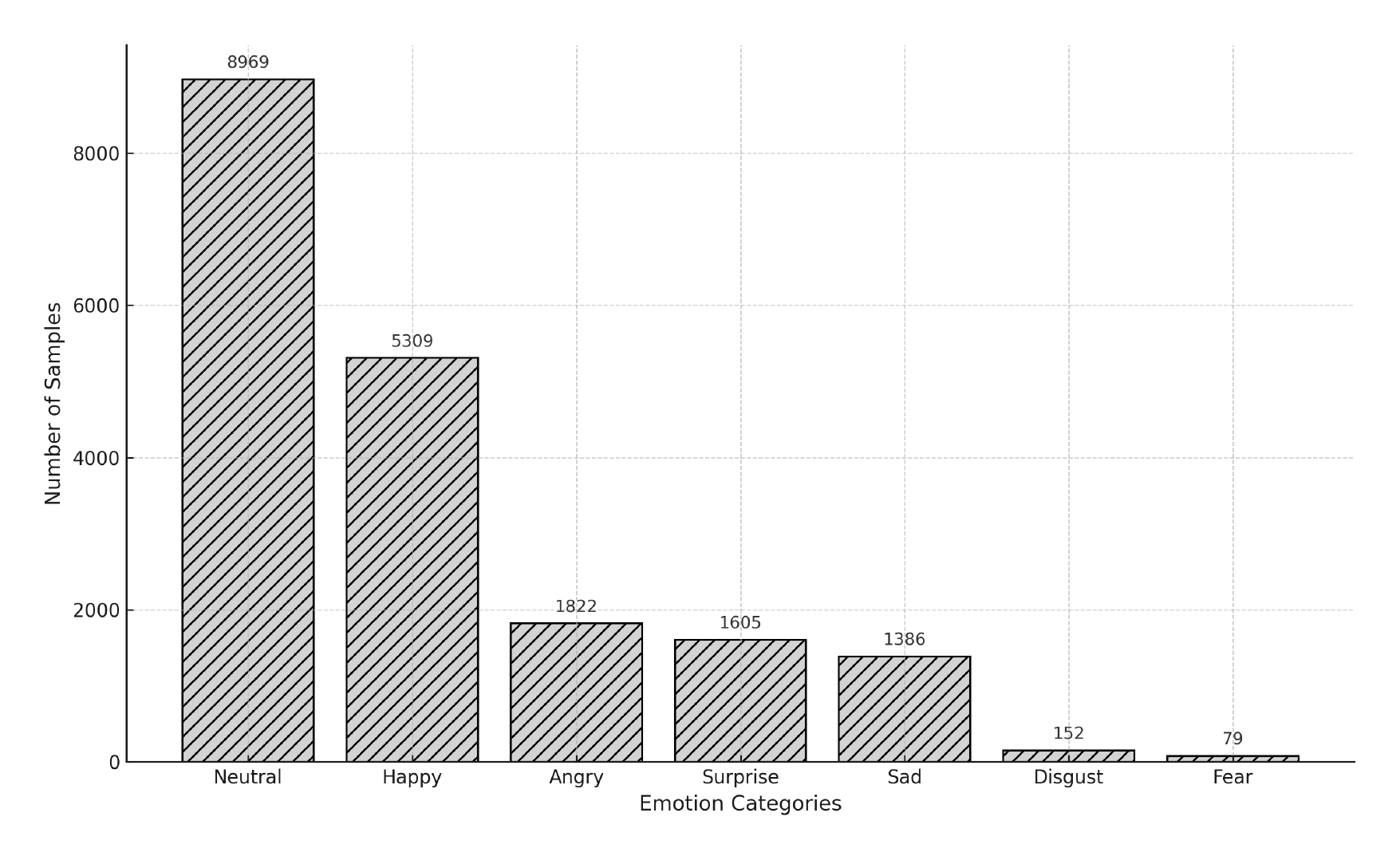}
\caption{Bar-chart representation of emotion distribution.}
\label{Figure 5}
\end{figure}
\vspace{-1.5 em}
\subsection{Prediction Analysis}

In order to quantitatively assess our ensemble-based emotion recognition system on responses of ASD children, a multi-layered visual and statistical analysis was conducted across seven emotion categories: \textit{happy}, \textit{sad}, \textit{angry}, \textit{fear}, \textit{disgust}, \textit{surprise}, and \textit{neutral}. Emotion-wise softmax scores of the Fusion-N model were investigated for prediction confidence, shape of distribution, and separability between classes. From \texttt{emotion\_descriptive\_stats.csv}, mean confidence values suggested \textit{happy} ($M=0.1459$), \textit{sad} ($M=0.1443$), and \textit{surprise} ($M=0.1434$) to be most prevailing, with \textit{neutral} lowest ($M=0.1386$). Low model uncertainty is indicated by narrow standard deviations for all classes ($\sigma \approx 0.001\text{--}0.003$). 
\begin{figure}[ht]
  \centering
  \includegraphics[width=0.45\textwidth]{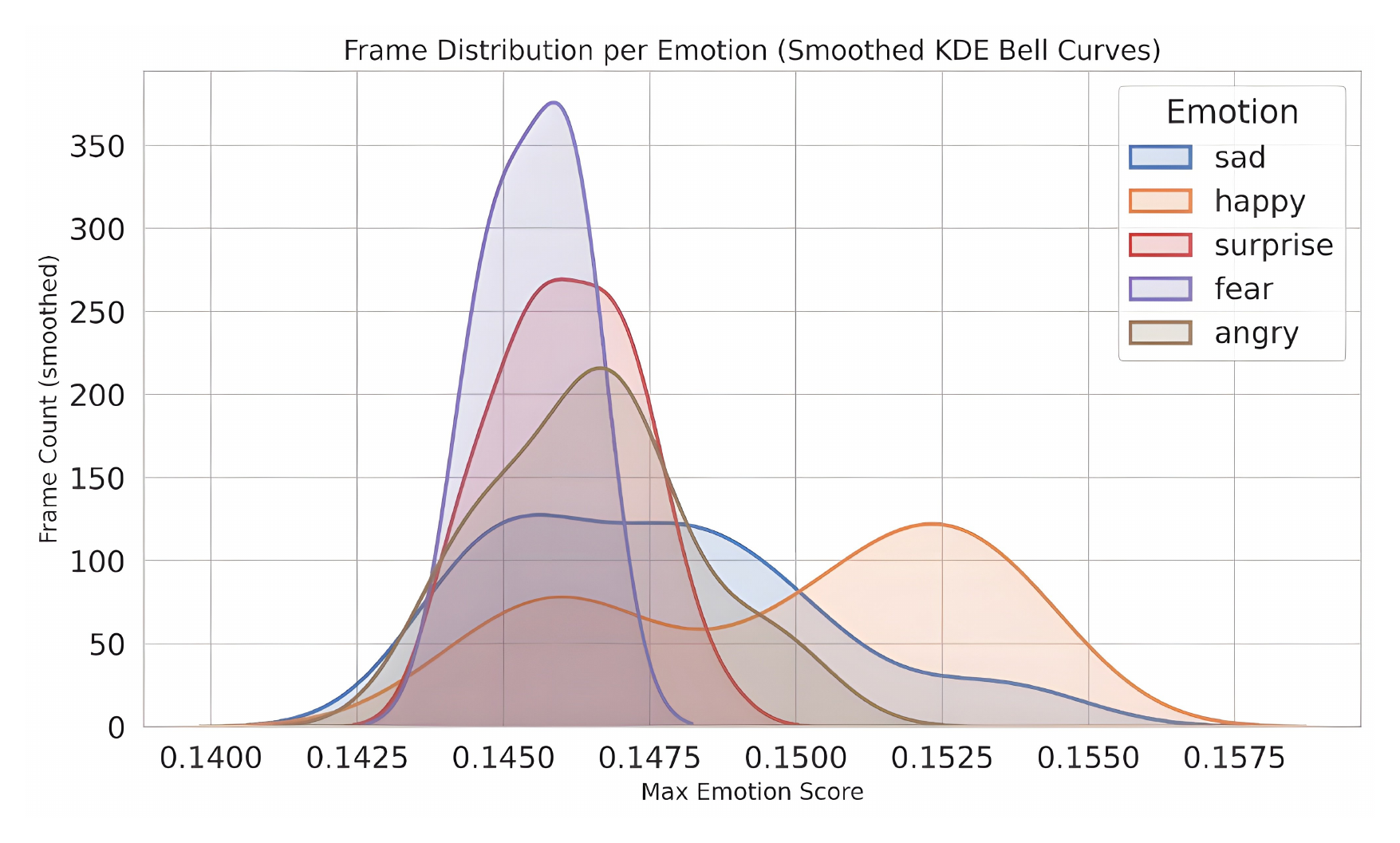}
  \caption{Smoothed KDE Curves for Emotion Scores.}
  \label{fig:kde}
\end{figure}

\begin{figure}[H]
  \centering
  \includegraphics[width=0.45\textwidth]{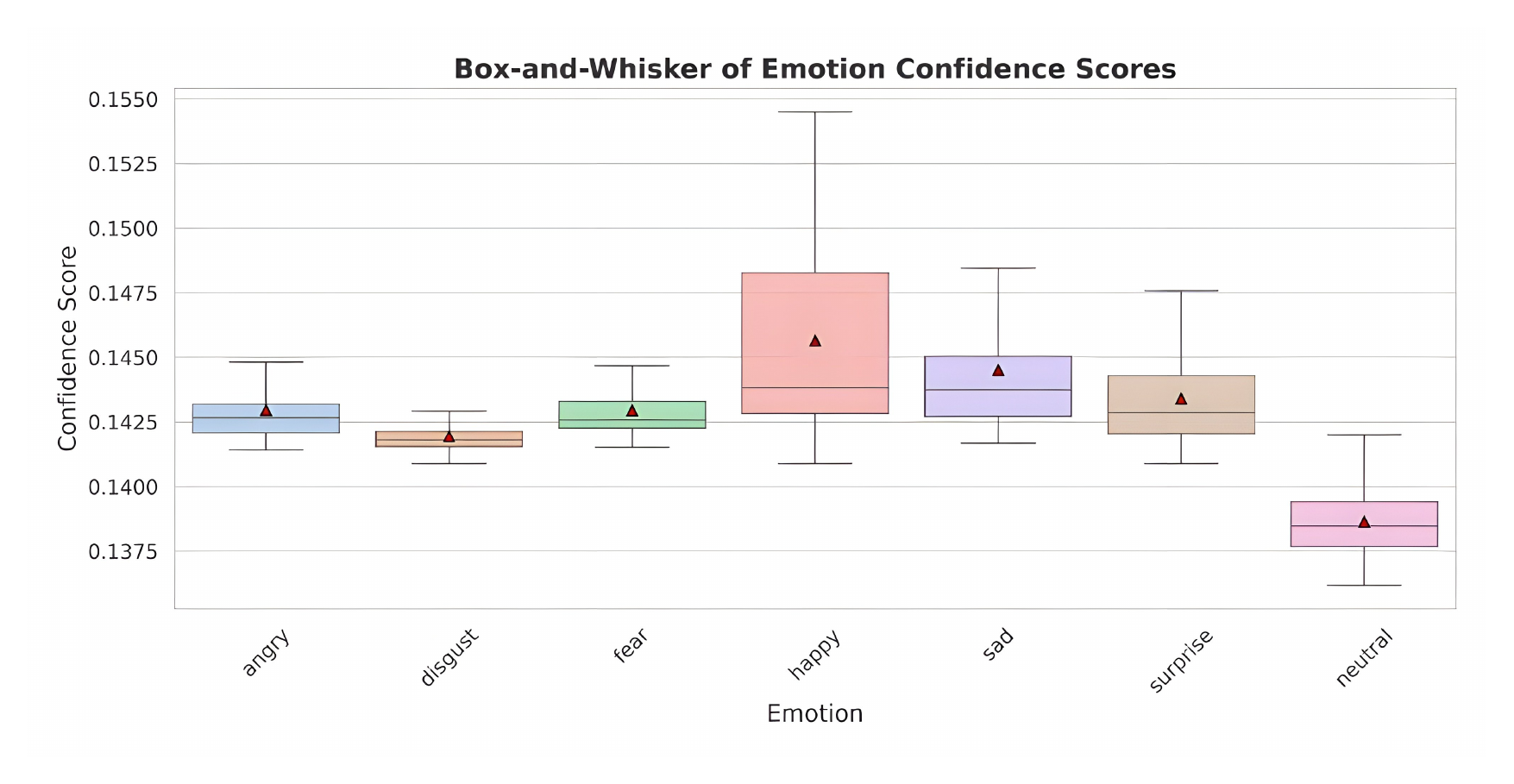}
  \caption{Box-Whisker Plot for Emotion Confidence Scores.}
  \label{fig:box}
\end{figure}

The boxplot (Fig.~\ref{fig:box}) indicated a greater median and wider outlier spread for \textit{happy}, tightly concentrated in $[0.145,0.155]$, while \textit{neutral} was tightly restricted in $[0.138,0.140]$. KDE smoothing  indicated (Fig.~9) a right-skewed peak for \textit{happy} ($\approx 0.148$), while overlapping distributions for \textit{sad}, \textit{fear}, and \textit{angry} reflect difficulties in distinguishing among these emotions due to their subtle expressivity in ASD.

Additionally, to examine the overall emotional tendencies of the autistic children, we classified the emotions that were observed during name-calling event into two categories : positive (happy, surprise) and negative (sad,angry,disgust). Fig 11 (pie-chart) shows that the majority of children, i.e, 73.3 \% (11 out of 15) exhibited predominantly positive emotions and the rest 26.7 \%(4 out of 15) were dominated by negative emotions. This observation aligns with prior work showing that robot-based interactive interventions can foster engagement and elicit positive responses in children with ASD \cite{Alarcon2021}.

\begin{figure}[ht]
  \centering
  \includegraphics[width=0.37\textwidth]{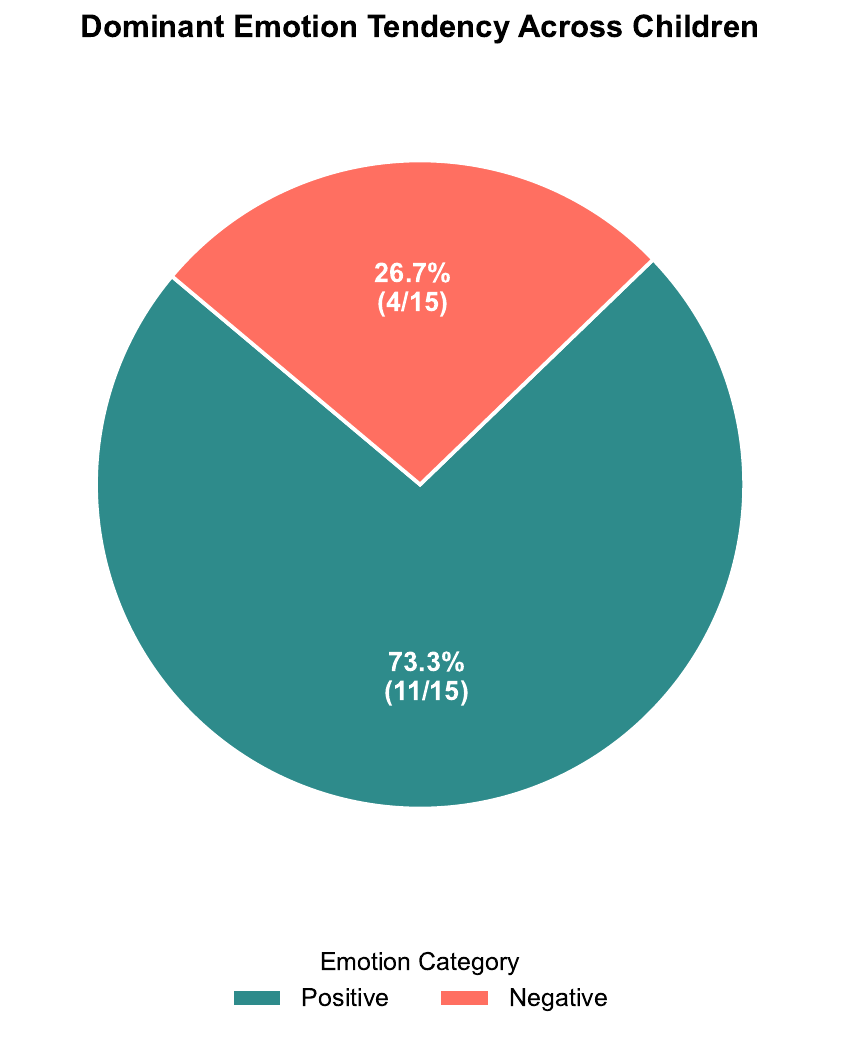}
  \caption{Pie-chart representing distribution of positive vs negative emotions on name-calling event. Teal shade represents positive (happy, surprise) emotions and coral shade represents negative emotions (sad, angry, disgust,fear).}
  \label{fig:kde}
\end{figure}

\subsection{Statistical Significance Testing}
ANOVA and Kruskal--Wallis tests between the seven emotion classes verified significant variation in model confidence:
\begin{itemize}
  \item \textbf{ANOVA:} $F(6, N) = 202.00$, $p < 1.0 \times 10^{-180}$
  \item \textbf{Kruskal--Wallis:} $H(6) = 692.18$, $p < 3.0 \times 10^{-146}$
\end{itemize}

Post-hoc Tukey HSD tests indicated that \textit{neutral} was always separable, with significantly lower confidence than \textit{happy}, \textit{sad}, \textit{angry}, and \textit{disgust} ($p < 0.001$). Both \textit{happy} and \textit{sad} achieved significantly higher confidence than \textit{neutral} and \textit{disgust}, demonstrating their salience in the ensemble's predictions.

\section{Conclusion}
\subsection{Ensemble-based labeling framework}
The proposed framework integrates predictions from pre-trained models (DeepFace's and FER) using a consensus strategy tailored for the expressive variability of autistic children. Given the inconsistent performance of off-the-shelf models on neurodiverse datasets, our ensemble was optimized to enhance robustness on ASD-specific facial data.

To assess generalizability, we evaluated the ensemble on a publicly available ASD dataset~\cite{talaat2023dataset} , annotated by a certified clinical psychologist. The model achieved 90.16\% accuracy relative to expert labels (Table~IV), demonstrating strong clinical concordance and adaptability to unseen data. Our results support ensemble learning as a scalable, clinically-aligned alternative to manual annotation in resource-constrained settings.

\subsection{Predictive hypothesis}
We compared emotion predictions made by 15 children with autism during human--robot interaction facilitated by the NAO robot comparing on 7 basic emotions. Descriptive statistics, visual distribution plots, and inferential statistical analyses were applied to determine emotional expressivity and inter-individual variability.

Mean and standard deviation values were calculated for each emotion per child. \textit{Happy}, \textit{sad} and \textit{surprise} exhibited higher mean scores across most participants, whereas \textit{neutral}, \textit{disgust}, and \textit{angry} remained at lower and relatively stable levels. Standard deviation patterns indicated greater variability in \textit{happy}, \textit{sad}, and \textit{fear}, while \textit{disgust} and \textit{neutral} were more consistent.

\textit{Participant-8}, \textit{Participant-9} and \textit{Participant-10} demonstrated a higher prevalence of \textit{happy} and \textit{sad} predictions, consistent with the theory of emotional salience in autism spectrum disorder (ASD)~\cite{Cassel2019}. The emotion \textit{fear} was more dominant in some children, reinforcing prior findings that ASD individuals often exhibit elevated anxiety or hyperarousal in novel contexts such as robot interaction~\cite{Costa2018}.

The emotions \textit{happy}, \textit{sad} and \textit{surprise} exhibited broader confidence intervals and denser distributions, suggesting their richer expressivity. The box-and-whisker plots confirmed this with larger inter-quartile ranges. There were several outliers as well in these emotions indicating transient emotional bursts, a known characteristic of affect dysregulation in ASD \cite{Macari2022}. This aligns with the known heterogeneity in affective displays among individuals on the autism spectrum, where emotional responses can range from subdued to highly exaggerated depending on context, sensory sensitivity, or individual traits.

\subsubsection*{Implications and Literature Alignment}

Our results are consistent with psychological research on emotion expression in ASD, where children with developmental or emotional difficulties possess an innate bias toward positive expressions in interactive and observational situations. In our dataset, 73.3 \% of the children exhibited a positive emotional dominance, represented by happy and surprise. An interesting minority (26.7\%), however, manifested a negative dominance, namely sad, disgust, and angry, seen among participants 2, 5, 6, and 7. This diversity highlights the importance of individualized, emotion-sensitive interventions since children with the overarching negative affect can be helped through specialized intervention in affective learning environments. Furthermore, these findings verify the viability of using robotic stimuli like NAO to examine and perhaps augment autistic children's emotional expressivity, and demonstrate the potential of emotion-aware robotics as a tool in affective computing and autism therapy.

\subsection{Future Scope and Discussions}

While the current system performs reliably in offline conditions, its application in real-time scenarios remains a key area for enhancement. As of now, NAO is being used only as a facilator, the primary limitation lies in latency introduced by sequential modules, particularly during face detection and preprocessing.

Future efforts can focus on optimizing the pipeline for real-time deployment by  prioritising low-latency, adaptive, and hardware-efficient implementations to extend real-world applicability.

Adaptive learning with reference to personal emotional profiles can improve performance across various ASD settings by detecting nuanced differences in affective expressions. Tested and validated using a geographically representative dataset, our ResNet-50 + three-layer GCN architecture presents strong, generalizable capability for ASD emotion analysis in real-world scenarios.
\section{Acknowledgement}
The authors thank the Smart Materials, Structures and Systems Laboratory of the Department of Mechanical Engineering and the Psychology Laboratory of the Department of Humanities and Social Sciences, IIT Kanpur, for the infrastructural facilities and support extended in conducting this research work. Special thanks go to Mr. Rohit Kumar Tiwari, a specialised clinical psychologist (Rehabilitation Psychology) at the Pushpa Khanna Memorial Centre, for his guidance in behavioral assessment, labelling of our global dataset, and support for dataset annotation. We also appreciate the cooperation and provision of logistic support from the Amrita Rehabilitation Centre and Pushpa Khanna Memorial Centre, both situated in Kanpur, India. Our sincere appreciation extends to the parents for trusting us and to the children for their voluntary participation. We lastly acknowledge all personnel who assisted in the process of data collection at partner centers and in our laboratory.

\bibliography{references}

\vfill

\end{document}